\documentclass[conference]{IEEEtran}
\IEEEoverridecommandlockouts
\usepackage{cite}
\usepackage{amsmath,amssymb,amsfonts}
\usepackage{algorithmic}
\usepackage{graphicx}
\usepackage{stfloats}
\usepackage{textcomp}
\usepackage{xcolor}
\usepackage{comment}
\usepackage{float}
\usepackage{xspace}
\usepackage{multirow}
\usepackage{tabularx}
\usepackage{enumitem}
\usepackage{hyperref}
\usepackage{float}
\usepackage{booktabs}
\newtheorem{definition}{Definition}
\newcommand{\SANA}{$\mathsf{{SANAGRAPH}}$\xspace}

\def\BibTeX{{\rm B\kern-.05em{\sc i\kern-.025em b}\kern-.08em
    T\kern-.1667em\lower.7ex\hbox{E}\kern-.125emX}}
\begin{document}

\title{Graph-based Event Log Repair 
}

\author{\IEEEauthorblockN{1\textsuperscript{st} Sebastiano Dissegna}
\IEEEauthorblockA{\textit{Department of Computer Science Engineering} \\
\textit{University of Trento}\\
Trento, Italy \\
sebastiano.dissegna@unitn.it}

\and

\IEEEauthorblockN{2\textsuperscript{nd} Chiara Di Francescomarino}
\IEEEauthorblockA{\textit{Department of Computer Science Engineering} \\
\textit{University of Trento}\\
Trento, Italy \\
c.difrancescomarino@unitn.it}
\and

\IEEEauthorblockN{3\textsuperscript{rd} Massimiliano Ronzani}
\IEEEauthorblockA{\textit{Fondazione Bruno Kessler}\\
Trento, Italy \\
mronzani@fbk.eu}

}

\maketitle

\begin{abstract}

The quality of event logs in Process Mining is crucial when applying any form of analysis to them. In real-world event logs, the acquisition of data can be non-trivial (e.g., due to the execution of manual activities and related manual recording or to issues in collecting, for each event, all its attributes), and often may end up with events recorded with some missing information. Standard approaches to the problem of trace (or log) reconstruction either require the availability of a process model that is used to fill missing values by leveraging different reasoning techniques or employ a Machine Learning/Deep Learning model to restore the missing values by learning from similar cases. 
In recent years, a new type of Deep Learning model that is capable of handling input data encoded as graphs has emerged, namely Graph Neural Networks. Graph Neural Network models, and even more so Heterogeneous Graph Neural Networks, offer the advantage of working with a more natural representation of complex multi-modal sequences like the execution traces in Process Mining, allowing for more expressive and semantically rich encodings. 

In this work, we focus on the development of a Heterogeneous Graph Neural Network model that, given a trace containing some incomplete events, will return the full set of attributes missing from those events. We evaluate our work against a state-of-the-art approach leveraging autoencoders on two synthetic logs and four real event logs, on different types of missing values. Different from state-of-the-art model-free approaches, which mainly focus on repairing a subset of event attributes, the proposed approach shows very good performance in reconstructing all different event attributes. 

\end{abstract}


\section{Introduction}
Process Mining (PM) is an interdisciplinary field that combines data science and process management, enabling organizations to analyze and improve their operational processes using data extracted from event logs. However, the reliability and accuracy of any PM  analysis fundamentally depend on the quality and completeness of the underlying event logs \cite{Log_quality, DBLP:books/sp/Aalst16}.
In practice, generating high-quality event logs is far from straightforward. Various challenges might arise due to the manual nature of certain activities, inconsistent data recording practices, or technical limitations during data acquisition. As a result, event logs often contain missing or incomplete information, especially related to individual event attributes. Hence the need for methods able to  \emph{repair event logs} and
reconstruct the missing pieces of a trace. 
Traditional approaches to event log (or trace) repair in case of missing data generally fall into two main categories. The first one relies on the availability of a process model, which is then used to infer missing values through logical or statistical reasoning. The second utilizes Machine Learning (ML) or Deep Learning (DL) techniques to learn patterns from complete cases and apply them to repair incomplete traces. While these methods have shown promise, they often struggle to capture the full complexity of inter-attribute dependencies. This holds since most approaches only focus on a subset of event attributes, disregarding secondary ones, which might also contain relevant pieces of information. 

Recently, Graph Neural Networks (GNNs) have gained attention \cite{wu_comprehensive_2021} for their ability to process data represented as graph structures that naturally capture the relationships and dependencies within the encoded trace attributes. In the context of PM, GNNs provide an expressive framework for representing execution traces, where events and their attributes can be modeled as nodes and edges in a graph\cite{Sephigraph, pasquadibisceglie_prophet_2024}.
In this work, we introduce \SANA, which aims to tackle the problem of repairing event logs with missing information, with a focus on cases where all the attributes of an event are missing. This is achieved by leveraging a semantically rich encoding as a heterogeneous graph \cite{wu2022gnn}, which, compared to its homogeneous counterpart, allows us to give a specific node type to each event attribute in the trace. This turns the repair problem into a node classification task, well-known in the GNN literature\cite{wu_comprehensive_2021}. The Heterogeneous GNN (HGNN) model allows for information to circulate across the graph in input and enrich the nodes marked as empty, which represents the missing information. A set of linear layers, one for each attribute, classify then the empty nodes into their respective repaired values. 
We evaluate our approach on two synthetic and four real-world event logs, with various missing data patterns. In contrast to existing state-of-the-art methods, most of which consider only a subset of missing attributes, our model demonstrates strong performance in reconstructing the complete set of attributes, showing its potential for enhancing data quality in PM applications.

The paper is structured as follows. In Section~\ref{sec:background}, the background theory on PM and GNNs is introduced. Section~\ref{sec:relWorks} contains an analysis of the current state-of-the-art approaches. In Section~\ref{sec:Encoding}, the heterogeneous graph encoding of traces, along with a description of the HGNN model employed and how it is trained is reported. Section~\ref{sec:settings} and Section~\ref{sec:results} discuss how the evaluation was carried out and the obtained results, respectively. Finally, Section~\ref{sec:results} concludes the paper and discusses future directions.

\section{Background}
In this section we discuss the background theory on Process Mining and Graph Neural Networks.
\label{sec:background}
\subsection{Process Mining}

In Process Mining \cite{van_der_aalst_process_2022,DBLP:books/sp/Aalst16}, process executions can be represented in event logs, which are a description of the events in the process of interest.  
\begin{definition}[Event Log]
        An event log $\mathcal{E}$ is a collection of events that correspond to the selected process. Each event $e \in \mathcal{E}$ has a series of attributes $a \in \mathcal{AT}$, where $\mathcal{AT}$ is the set of all possible attributes.
        The three main attributes are as follows: 
\begin{itemize}[label=$\bullet$]
    \item $Case_{ID}$ ($e^{id}$): this identifies the unique case a specific event belongs to;
    \item activity ($e^{act} \in \mathcal{A}$): this is the activity (in the set of activities $\mathcal{A}$) performed in that event;
    \item timestamp ($e^{ts}$): this indicates the time at which the event occurred.
\end{itemize}
\end{definition}

\begin{table*}[t]
    \centering
    \scalebox{0.85}{
    \begin{tabular}{c@{\hskip 0.1in}  c@{\hskip 0.1in}  c@{\hskip 0.1in}  c@{\hskip 0.1in} c@{\hskip 0.1in}  c @{\hskip 0.1in}  c }
        \toprule
        \textbf{CaseID}  & \textbf{Activity} & \textbf{time:timestamp} & \textbf{org:resource} & \textbf{lifecycle:transition} & \textbf{case:REG\_DATE} &  \textbf{case:AMOUNT\_REQ} \\
        \toprule
         173688 &A\_SUBMITTED & 2011-09-30 22:38:44.546000+00:00 &112 & COMPLETE & 2011-10-01 00:38:44.546000+02:00 & 20000 \\
        173688 & A\_PARTLYSUBMITTED & 2011-09-30 22:38:44.880000+00:00 &112 & COMPLETE & 2011-10-01 00:38:44.546000+02:00 & 20000 \\
        173688 & A\_PREACCEPTED & 2011-09-30 22:39:37.906000+00:00 &112 & COMPLETE & 2011-10-01 00:38:44.546000+02:00 & 20000\\
        173688 & A\_ACCEPTED & 2011-10-01 09:42:43.308000+00:00 &10862 & COMPLETE & 2011-10-01 00:38:44.546000+02:00& 20000 \\
        173688 &A\_FINALIZED & 2011-10-01 09:45:09.243000+00:00 &10862 &COMPLETE &2011-10-01 00:38:44.546000+02:00 & 20000 \\
        \hline
    \end{tabular}
    }
    \vspace*{1mm}
    \caption{Example of a trace belonging to the BPI 2012 dataset}
    \vspace{-0.4cm}
    \label{tab:trace_example}
\end{table*}

\begin{table*}[t]
    \centering
    \scalebox{0.85}{
    \begin{tabular}{c@{\hskip 0.1in}  c@{\hskip 0.1in}  c@{\hskip 0.1in}  c@{\hskip 0.1in} c@{\hskip 0.1in}  c @{\hskip 0.1in}  c }
        \toprule
        \textbf{CaseID}  & \textbf{Activity} & \textbf{time:timestamp} & \textbf{org:resource} & \textbf{lifecycle:transition} & \textbf{case:REG\_DATE} &  \textbf{case:AMOUNT\_REQ} \\
        \toprule
         173688 &- & 2011-09-30 22:38:44.546000+00:00 &112 & COMPLETE & 2011-10-01 00:38:44.546000+02:00 & 20000 \\
        173688 & A\_PARTLYSUBMITTED & - &112 & COMPLETE & 2011-10-01 00:38:44.546000+02:00 & 20000 \\
        173688 & - & - & - & - & - & -\\
        173688 & A\_ACCEPTED & 2011-10-01 09:42:43.308000+00:00 &10862 & COMPLETE & 2011-10-01 00:38:44.546000+02:00& 20000 \\
        173688 & - & 2011-10-01 09:45:09.243000+00:00 &10862 & COMPLETE & 2011-10-01 00:38:44.546000+02:00& 20000 \\
        \hline
    \end{tabular}
    }
    \vspace*{1mm}
    \caption{Example of a trace with missing attributes and events}
    \vspace{-0.8cm}
    \label{tab:trace_example_missing}
\end{table*}

Along with the three aforementioned main attributes, each event may contain additional (categorical or numerical) attributes. Examples of attributes belonging to an event include the resource that has carried out the activity, e.g., the person responsible for it;  the cost of executing a certain event; or information about how much time has elapsed since the start of the process or since the previous event occurred.

An event log is a collection of traces that are identified via the $Case_{ID}$ attribute, which is unique to that case. 
\begin{definition}[Trace]
    A trace $\sigma \models \langle e_1, .., e_{|\sigma|}\rangle$ is a non-empty sequence of events $\sigma \in \mathcal{E}^*\backslash\{\emptyset\}$, such that $\forall e_i,e_j \in \sigma, 1 \leq i \leq j \leq |\sigma| : e_i^{id} = e_j^{id} \land e_i^{ts} \le e_j^{ts}$, i.e., they all have the same $Case_{ID}$ and are ordered according to the timestamps.
\end{definition}

 We distinguish between \textit{trace attributes} (static attributes), whose value is constant throughout the whole trace, and \textit{event attributes} (dynamic attributes), whose value can change among different events of a trace. 
In Table \ref{tab:trace_example}, we report an example of a trace contained within the $BPI 2012$ \cite{BPI_12} event log. This particular event log contains information about the loan application process, from submission to approval or rejection, of a Dutch financial institution. In the example, for instance, \textit{case:AMOUNT\_REQ} is a trace attribute and \textit{org:resource} is an event attribute.
Event logs may contain missing information that hampers the analysis of the data. For example, Table \ref{tab:trace_example_missing} shows how the trace $173688$ might appear with missing information: the third event is entirely absent, the \textit{activity} attribute is missing from the first and last events, and the second event lacks a timestamp. In order to be used in PM algorithms, its missing values would need to be reconstructed.

\subsection{Graph Neural Networks}
Compared to other DL architectures GNNs are specifically built to work with data that are represented as graphs. A graph \textit{G} is defined via a tuple \textit{(V, E)}, where $V$ is the set of nodes (vertices), and $E$ is the set of edges (arcs). With $v_i \in V$ we denote a node and with $e_{ij} \in E$ we denote the edge that goes from node $v_i$ to node $v_j$. 
We can represent the connectivity of a graph with a corresponding adjacency matrix $A$. An adjacency matrix is a matrix of boolean values that has dimension $n \times n$, where $n$ is the number of nodes. It follows that, if $A_{ij} = True$ then there exists an edge $e_{ij} \in E$ connecting node $v_i$ and node $v_j$, $A_{ij} = False$ otherwise. In conjunction with the adjacency matrix, we need two more matrices to represent the graph features. The first matrix, $X^V \in \mathbb{R}^{n \times d}$, represents the node features, where each row $x_v \in \mathbb{R}^d$ corresponds to a node and $d$ denotes the dimensionality of its feature vector. The second one, which is optional in many architectures, is the edge attribute matrix $X^E$, where $X^E \in \mathbb{R}^{m \times c}$, with $x^e_{ij}$ representing an edge feature vector, $m$ the number of edges, and $c$ the dimensionality of each edge vector. The neighborhood of a node $v$ is defined as $N(v) = \{u \in V | e_{vu} \in E\}$, which is the set of nodes such that there exists an edge from $v$ to them. 

GNNs can be used to address three different types of problems. 
The first one regards \textit{node level tasks}, that is, node classification or node regression. With a GNN, we can indeed obtain high-level node representations, which can be fed to a Multi-Layer Perceptron (MLP) to predict a node class in an end-to-end pipeline. The second type deals with \textit{edge level tasks}, namely edge classification or link prediction, with the latter predicting the presence or not of an edge between two nodes. The last type concerns \textit{graph classification tasks}, which summarize the information contained in the entire graph in input and classify it according to the specific class. 

Moreover, the graphs provided as input to a GNN can be of two types, \textit{homogeneous} or \textit{heterogeneous}. In the former case, all nodes in the graph are of the same type, and it follows that their node feature vectors contain the same amount and type of information, i.e., they have the same dimensionality. In the latter case, nodes and edges can be of different types.

Convolutional GNN (CGNN) is one of the most used types of GNNs. By leveraging the developments that Convolutional Neural Networks obtained in the Computer Vision field, these approaches extend the convolution operation to graphs.
Similarly to what happens in a convolutional step in image processing, where a pixel representation is updated according to the value of the other pixels inside the filter's window, in a CGNN, the node representation is updated according to the value of the nodes inside its neighborhood.
An advantage of CGNNs compared to the 2D convolutional operation lies in the fact that there is no fixed size of the node's neighborhood as in the pixel case, thus removing the need to employ zero padding by adding fake nodes, which is what happens when working with images, where pixels with values set to zero are needed for the operation to work correctly.

From CGNNs, two different branches of approaches are being developed, namely, \textit{Spectral-based approaches} and \textit{Spatial-based approaches}. We are interested in the Spatial-based family of approaches.

\subsubsection{Spatial-based Convolutional Graph Neural Network}

This family of architectures defines the graph convolutional operation based on a node's local neighborhood.
In general, a node’s representation is updated by considering both its representation from the previous layer and the representations of its neighboring nodes. When applied to the problem of graph classification, graph pooling layers can be employed with this family of architectures. These layers are responsible for reducing the graph dimensionality by aggregating groups of nodes into a single one. This helps make the end-to-end training process more efficient.

The Message Passing Neural Network MPNN is a general spatial-based model upon which more refined architectures can be built to address specific problems. With this model, we perform a K-step message-passing operation to propagate the information across the entire graph, where $K$ is the number of layers. During the message passing phase, hidden states $h_v^{k}$ at each node $v$ in the graph are updated as follows 
\begin{equation}
\label{eq:message_passing}
    h_v^{k} = UP_k \left( h_v^{k-1}, \sum_{u\in N(v)} AGRR_k \left( h_v^{k-1}, h_u^{k-1}, x^e_{vu}\right) \right).
\end{equation}

By looking at Equation~\eqref{eq:message_passing}, we can notice two functions. The first one, $AGGR$, is responsible for aggregating the information obtained by taking as input the representation at the previous layer of the node we are updating, along with the representation of each neighbor and the edge feature vector that connects them. As stated before, the feature vector on the edge is optional. The second function 
is an update function, $UP$, which is applied to the sum of the message information from the neighborhood and to the previous node representation. Both $AGRR$ and $UP$ have learnable parameters.

Finally, if we are performing graph classification, we apply a $READOUT$ function, 
which also has learnable parameters and  summarizes the information contained in the entire graph
\begin{equation}
\label{eq:readout}
    h_G = READOUT \left( h_v^K \mid v\in V\right).
\end{equation}

\paragraph{Graph Attention Network (GAT)}
All the aforementioned methods assume that the contribution of neighboring nodes is either identical or predetermined. In the GAT architecture \cite{GAT}, an attention mechanism, described in Equation~\ref{eq:gat}, is introduced to make the weights between two connected nodes learnable:
\begin{equation}
\label{eq:gat}
    h_v^k = \zeta \left( \sum_{u\in N(v) \cup v} \alpha^k_{vu} W^k h_u^{k-1} \right)
\end{equation} 
with $\zeta$ a non-linear function.
The attention coefficient $\alpha^k_{vu}$, which represents the importance of node $u$ to node $v$ at layer $k$, is learned via an end-to-end neural network architecture. This mechanism assigns higher weights to more important nodes, as described by the formula:
\begin{equation}
\label{eq:attention}
    \alpha^k_{vu} = SOFTMAX(LeakyReLU(a^T[W^kh_v^{k-1}||W^kh^{k-1}_u]))
\end{equation}
where $a$ and $W$ are learnable parameters, and $||$ denotes the concatenation operator.

\paragraph{SAGE Convolution} 
For this project, we rely on the simple convolution operator detailed in \cite{GRAPHSAGE}. In which, following~\eqref{eq:SAGEConv}, we update a node representation by adding the sum between the product of the current node representation $x_i$ and two sets of learnable parameters $(W_1, W_2)$ and the mean aggregation of the node neighborhood.
\begin{equation}
    \label{eq:SAGEConv}
    x_i^\prime = W_1x_i + W_2 \cdot \text{mean}_{j\in N(i)}x_j 
\end{equation}

We use this simple operator primarily because, in our problem's encoding, much of the input graph lacks information due to the presence of empty nodes that need to be restored. This sparsity makes the use of the graph attention operator \cite{GAT} impractical. 

\section{Related Work}
\label{sec:relWorks}

Omissions in event log data -- i.e., relevant information that is not recorded -- are recognized as a primary factor contributing to the deterioration of event log quality \cite{Log_quality}.
Consequently, many works have addressed the problem of missing information in event logs, with a particular focus on trace completion. Existing approaches can be broadly categorized into two main classes: those that rely on a-priori knowledge (e.g., process models) and those that operate without such information. This section provides a brief overview of the most relevant contributions, organized according to this distinction.

\emph{Model-based approaches} have emerged as a natural extension of conformance checking in Process Mining \cite{handbook_conformal_checking}, particularly through the application of alignment techniques \cite{alignment}, where incomplete traces are repaired by aligning them to the closest behavior allowed by the process model.
These techniques can be extended to support the reconstruction of data attributes by leveraging data-aware conformance checking \cite{DA_conformal_checking}.
In \cite{bayesian_nets}, the authors propose a method to recover missing information about executed activities and their durations by combining alignment algorithms with Bayesian networks.

A distinct model representation is used in \cite{causal_net_missing_event}, where the authors combine causal networks with a branching framework to repair missing activities. The efficiency challenges encountered by this work are tackled in 
\cite{process_tree_repair}, which uses process tree model decomposition instead of Petri nets to represent the process, as the latter often leads to inaccurate activity reconstructions in scenarios involving complex and nested loop structures. Moreover, introducing an innovative branch indexing technique, the authors mitigate the problem of state space explosion.

The works in \cite{Chiara_CAISE13, DeMasellis_BPMW2018} adopt a different approach: data-enriched workflow nets are used to model the process. The trace completion task is then reformulated as a reachability problem, which can be solved using planning techniques. 

Finally, unlike previous works,\cite{missing_case_id} addresses the retrieval of case identifiers in unlabeled event logs by exploiting the structure of an existing process model to infer the correct case associations.

\emph{Model-free approaches} are less common but offer broader applicability, as they work directly from event data without requiring any a-priori knowledge such as a process model.  This work falls in this second category of approaches. 

Most of the existing approaches aim at repairing missing activities. 
For instance, \cite{succession_relations} presents a method for reconstructing missing activities based on activity inheritance relationships in event logs. Incomplete traces are completed using information from complete traces within the most similar cluster, identified through distance measures and trace frequency. \cite{Lu2022_LSTM} leverages Long-Short Term Memory networks to predict missing activity labels.
 \cite{PingWu} adopts a Transformer-based architecture to retrieve missing activities and correct anomalies, leveraging contextual dependencies within traces. 


To the best of our knowledge, only \cite{Comuzzi_Autoencoders} addresses the repair of missing data beyond activity labels.  
Building on the idea of approaches employing  an autoencoder architecture to recover missing information in electronic health records \cite{AE_health_records}, \cite{Comuzzi_Autoencoders} uses autoencoders to exploit complex relationships among attribute values in order to reconstruct missing data and correct erroneous entries.
Unlike all these works, \SANA aims at repairing different types of missing values. We hence compare it with \cite{Comuzzi_Autoencoders} in the evaluation, as it is the closest in terms of types of attribute labels it is able to reconstruct.

\section{Graph-based Trace Encoding and Model overview}
In this section we present our graph-based solution \SANA. Specifically, we first discuss the heterogeneous graph encoding of the traces extracted from the event logs, and then we introduce the architecture of the HGNN model employed.
\label{sec:Encoding}
\subsection{Trace to Graph}
In order to work with an HGNN model, we first encode  trace and event attributes into numerical vectors, and then create a graph deciding how many types of nodes we need and how to connect them. Starting from the node features, for categorical attributes we rely on the one-hot encoding, which converts a class label into a vector of length equal to the number of classes of that attribute and assign to each of them an index. We hence obtain a vector where one value is set to one, the class index, and all the others are zeros. Regarding numerical attributes, we apply a log-normalization, that is we apply the logarithmic function to all values, in order to reduce the variance of data for that specific attribute, and hence help the model converge faster. 
In the end, we have a node type for each attribute inside a trace. Moreover, we need to represent missing events. To do this, we add a class labeled \textit{MISSING VALUE} to all categorical attributes, while for numerical ones we use the value $-1$, which replaces the original value inside the trace for the missing events. 

\begin{figure}[t]
    \centering
    \scalebox{1}{
    \includegraphics[width=1\linewidth]{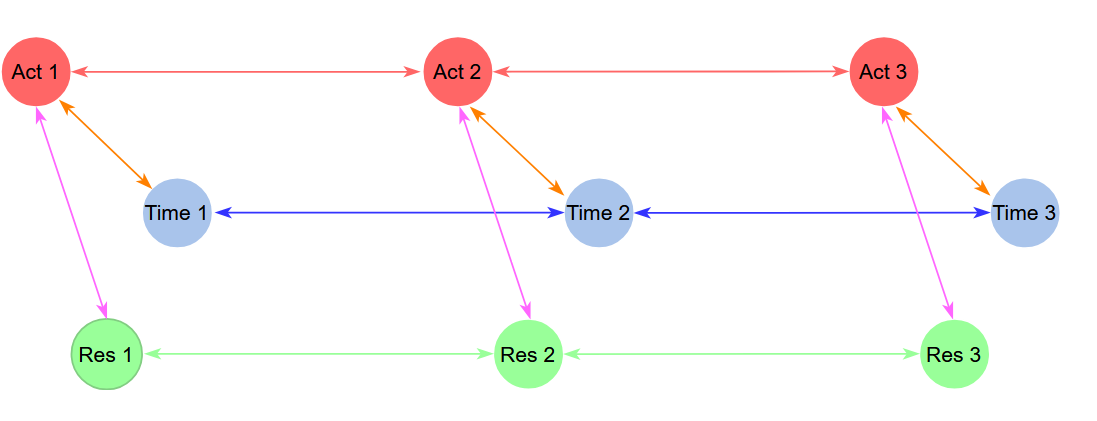}
    }
    \caption{Simple example of a heterogeneous graph encoding of a trace}
    \vspace{-0.5cm}
    \label{fig:trace_hgnn}
\end{figure}

Now that we have decided how to transform trace attributes into numerical vectors, we can create our heterogeneous graph.
We will have a node for each attribute of each event at a certain prefix length. So for a trace of length $m$ with $n$ attributes, we obtain $m\times n$ nodes. As for the connectivity, following the sequential order of the events in the trace according to the timestamps, node of the same type will be connected together, with the exception of the activities nodes, which will also be connected to the other node types, allowing for information to spread uniformly across the entire graph.  Figure~\ref{fig:trace_hgnn} shows an example of the (heterogeneous graph) encoding of a complete trace composed of 3 events containing the activity, timestamp, and resource attributes. 

\subsection{HGNN model}

In our model, we opt for the convolutional operator SAGEConv ~\cite{GRAPHSAGE}, which, in our initial experiments, demonstrated to be faster though with performance similar to the GAT ~\cite{GAT} mechanism. We attribute this to the fact that the input graphs contain many missing values, which likely reduce the effectiveness of the attention mechanism.
Unlike homogeneous GNNs, in  heterogeneous settings, like ours, we need to define one operator, which has its own set of parameters, for each type of edge in the encoded graph. Following the example in Figure~\ref{fig:trace_hgnn}, we will have the following edge types:

\begin{itemize}[topsep=0.1cm]
    \item[$\bullet$] $Activity_i \rightarrow Activity_{i+1}$
    \item[$\bullet$] $Activity_i \rightarrow  Time_i$
    \item[$\bullet$] $Activity_i \rightarrow  Resource_i$
    \item[$\bullet$] $Time_i \rightarrow  Time_{i+1}$
    \item[$\bullet$] $Resource_i \rightarrow  Resource_{i+1}$
\end{itemize}

It is also possible to choose a different type of operator for each edge type, but in our case, we simply use a SAGEConv type operator for all of them, described by Equation~\eqref{eq:SAGEConv}. This operator also does not allow for edge features, but this is not a problem since we don't have any edge features in this specific encoding. The choice of not employing any form of edge features is for the same reason as we do not employ the GAT layer, since most of the nodes are \textit{empty} we lack meaningful information about the relationships between nodes with known values and those with unknown values—apart from the existence of an \textit{empty} node.

The goal of the model is to assign to each node labeled as empty the correct class or the correct numerical value, for categorical and numerical nodes respectively. To achieve this, we use a mask associated with each input graph.
This mask is a boolean vector, with a length equal to the length of the corresponding trace, where \textit{TRUE} values indicate the empty events that require classification. 

We can retrieve the classification of the empty node by feeding after some convolutional layers its updated feature vector to a linear layer. The layer is then followed by a softmax in the case of categorical attributes, or its raw output is directly used in the case of numerical attributes. 

This model can be trained in an end-to-end manner by minimizing the following Loss Function, which combines a sum of $CrossEntropy$ losses over the categorical attributes with a sum of $L_1$ losses for the numerical ones as follows
\begin{equation}
    LOSS = \sum_{k \in \mathcal{C}_c}CrossEntropy(x_k, y_k) + \sum_{k \in \mathcal{C}_T}L_1(x_k,y_k),
    \label{eq:Loss function}
\end{equation}
where $x_k$ and $y_k$ are respectively the predicted class or numerical value and the ground truth for attribute $k$, and $L_1(x_k, y_k) = \frac{1}{n}|x_k - y_k|$ is the Mean Absolute Error Loss, and $n$ is the number of node to reconstruct.

\subsection{Number of convolutional layers}
In this model, the number of convolutional layers is a very critical parameter and deserves a bit of discussion. Compared to the graph classification problem in which we can compute a readout function to obtain a summary of all the information contained in the graph in input, when classifying a node, we only have access to the current hidden representation of the node after the convolutional layers. Let's assume we have a trace of 10 events and two layers of convolution. The fifth node will receive information from nodes in a window of 2 steps from its position, so from nodes (3,4,6,7) since edges are bidirectional. Now, if events at those 4 positions were also empty, then our fifth node would not have received any useful information that would allow the model to determine its correct class/value. Basically, this problem arises when there is a sequence of empty nodes two times larger than the number of convolutional layers. 
This is a very pessimistic scenario, and certain experiments in the evaluation are intentionally created to examine this limitation.
A possible solution to this problem is straightforward; we simply need to increase the number of layers, which comes at the cost of longer training times, as we will see in one of the experiments in Section \ref{sec:results}.

\section{Evaluation Setting and Implementation}

\label{sec:settings}

In this section, we present the research questions, the datasets, implementation details and the evaluation approach.

\subsection{Research questions}

In this work, we address the following research questions concerning the effectiveness of \SANA in repairing missing values in event logs:
\begin{enumerate}[label=\textbf{RQ\arabic*},leftmargin=1cm, topsep=0.1cm]
\item\label{RQ1} How does \SANA compare with state-of-the-art in terms of reconstructing missing activity labels and timestamps when only activity and timestamp information is considered?
\item\label{RQ2} How does using \SANA, for reconstructing all attribute information, impact its performance on activity and timestamp reconstruction?
\end{enumerate}

The first research question aims to assess the effectiveness of our method in repairing missing activity and timestamp information---a task that is within the capabilities of current state-of-the-art approaches.
The second research question aims to assess the impact of considering all attribute information in the reconstruction task. While the model can leverage contextual information from non-missing attributes, the task becomes more challenging, as it must handle the reconstruction of  multiple missing attribute values simultaneously.


In order to answer \textbf{RQ1}, we consider as main state-of-the-art approach the work presented in \cite{Comuzzi_Autoencoders}, which is based on deep learning autoencoder architectures. This method has proven to be the most effective existing solution for repairing missing event information that involves both activity and timestamp values. However, it is not designed to handle additional attributes.
%
To answer \textbf{RQ2}, we hence compare the activity and timestamp reconstruction results obtained by \SANA with all attributes with those obtained by \SANA with only activity labels and timestamps.

\subsection{Datasets}

The evaluation is carried out on 2 synthetic event logs and 4 real-life event logs available on the \href{https://data.4tu.nl/portal}{4TU Centre for Research} \footnote{https://data.4tu.nl/portal}, except for SP2020 that is available on \href{https://zenodo.org/records/3928487}{Zenodo} \footnote{https://zenodo.org/records/3928487}. The two synthetic datasets and BPI2012 and BPI2013 are the same datasets used in \cite{Comuzzi_Autoencoders}.

\begin{itemize}[label=$\bullet$,topsep=0.1cm]
    \item \textbf{Small and Large Log}: synthetic datasets used in \cite{Comuzzi_Autoencoders}
  
    \item \textbf{BPI-2012} \cite{BPI_12}: an event log of a loan application process.

    \item \textbf{BPI-2013} \cite{bpi13}: an event log containing traces of the problem management system of Volvo IT located in Belgium. 

    \item \textbf{BPI-2020-R} \cite{bpi20}: an event log containing traces related to Requests for Payment on two years of travel expense claims.
    \item \textbf{SP-2020} \cite{sp2020}: an event log from a service process of a home appliances vendor for repairing faulty devices.

\end{itemize}

The main characteristics of the datasets are reported in Table~\ref{tab:datasets}.

\begin{table}[t]
\scalebox{0.85}{
    \centering
\begin{tabular}{ l c@{\hskip 0.1in} c@{\hskip 0.1in} c@{\hskip 0.1in} c@{\hskip 0.1in} c@{\hskip 0.1in} c }
    \toprule
    & \textbf{Small Log} & \textbf{Large Log} & \textbf{BPI-2012} & \textbf{BPI-2013} & \textbf{BPI-2020-R} & \textbf{SP2020}\\
    \midrule
    trace \# & 2000 & 15000 & 13087 & 1487 & 6886 & 23906\\ \midrule
    event \# & 28000 & 120000 & 262200 & 6660 & 36796 & 178078\\ \midrule
    attribute \# & 2 & 2 & 6 & 12 & 7 & 5 \\ \bottomrule
\end{tabular}
    }
    \vspace*{1mm}
    \caption{Dataset information}
    \vspace{-0.8cm}
    \label{tab:datasets}
\end{table}

\subsection{Environment and Procedure}
\SANA has been developed on Python 3.9, using PyTorch 2.1 and PyTorch Geometric with CUDA 12.7. The hyperparameter optimization has been carried out using the Adaptive Experimentation platform.\footnote{\href{https://ax.dev/}{https://ax.dev/}} 
%
All experiments have been performed on a laptop equipped with an Intel (R) Core (TM) i7-13700H, 32 GB of RAM, and a 4070 Nvidia GPU. 

The hyperparameter optimization, whose objective is to minimize the total loss as detailed in (\ref{eq:Loss function}), has been conducted on the following set of parameters: \textit{Learning rate} in the range $[10^{-4}, 10^{-1}]$ on a logarithmic scale, \textit{Batch size} with values $\{16, 64, 256,512,1024,2048\}$, \textit{weight decay} with values in the range $[10^{-2}, 10^{-1}]$, \textit{Aggregation type} with values \{sum, mean, max\}, and \textit{Random seed} set to 123.
The total number of optimization trials has been set to 20 for bigger datasets and to 40 for smaller ones. The hidden size of our nodes has been set to 128 for all datasets, since it showed faster training times. The maximum number of epochs has been set to 50, with early stopping when the validation loss does not decrease for 5 consecutive epochs. 
We each dataset in $(60\%/20\%/20\%)$ for training, validation, and test sets. It is important to consider that when we are removing an event, we are removing it along with all of its attributes. This represents a worst-case scenario in event log repair, where usually only some attributes of an event are missing from a trace, not entire events.

In our experimentation, in order to test different types of incomplete data, we decided to create four types of masks to apply to a trace: 
\begin{itemize}
    \item \textit{ODD}, in which we remove all events whose index is an odd number inside a trace
    \item \textit{EVEN}, in which we remove all events whose index is an even number inside a trace
    \item \textit{WINDOW}, in which we leave an event and then remove the following two throughout the entire trace
    \item \textit{RANDOM}, in which we remove random events from a trace with a probability of $50\%$
\end{itemize}
It follows that from a trace we obtain four different types of partial traces. 

Our model is trained on all 4 masking strategies at the same time. For the testing of the performance, we take the best configuration of hyperparameters according to the search, and we use it to train 10 different models, which are tested on the 4 different dataset variants (one per masking strategy), separately. This allows us to check whether the model performance is stable by computing the mean and standard deviation for all the measures in the evaluation. Regarding the measures used, we have the accuracy for categorical attributes and the Mean Absolute Error (MAE) for numerical attributes. 

\section{Evaluation results}
In this section we report and discuss the results obtained with \SANA in the trace reconstruction task.\footnote{The code with the implementation of \SANA and all the evaluation results are available at \url{https://github.com/sebdisdv/SANAGRAPH}} 
\label{sec:results}

\subsection{Answering the Research Questions}

\begin{table*}[t]
\scalebox{0.93}{
    \centering
 \begin{tabular}{llcccccccccccc}
    \toprule
    \multirow{3}{*}{\textbf{Masking}} & \multirow{3}{*}{\textbf{Method}} & \multicolumn{2}{c}{\textbf{Small Log}} & \multicolumn{2}{c}{\textbf{Large Log}} & \multicolumn{2}{c}{\textbf{BPI12}} & \multicolumn{2}{c}{\textbf{BPI13}} & \multicolumn{2}{c}{\textbf{SP2020}} & \multicolumn{2}{c}{\textbf{BPI20RFP}}\\
    \cmidrule(lr){3-4} \cmidrule(lr){5-6} \cmidrule(lr){7-8} \cmidrule(lr){9-10} \cmidrule(lr){11-12} \cmidrule(lr){13-14}
    & & Activity & Time & Activity & Time & Activity & Time & Activity & Time & Activity & Time & Activity & Time \\
    \toprule
    \multirow{2}{*}{\textit{ODD}}
    & AE \cite{Comuzzi_Autoencoders} & 19.48 & 0.211 & 37.29 & 0.408 & 17.12 & \textbf{0.0025} & \textbf{64.65} & \textbf{0.003} & 35.22 & 0.048 & 56.94 & \textbf{0.004}\\
    & \SANA(AT) & \textbf{75.7} & \textbf{0.042} & \textbf{83.43} & \textbf{0.013} & \textbf{84.56} & 0.0172 & 45.32 & 0.014 & \textbf{61.7} & \textbf{0.012} & \textbf{98.59} & 0.02\\
    \midrule
    \multirow{2}{*}{\textit{EVEN}}
    &AE \cite{Comuzzi_Autoencoders} & 11.53 & 1.358 & 25 & 2.4166 & 16.07& \textbf{0.0037}& 34.15 & \textbf{0.0032} & 29.46 & 0.0746 & 61.63 &\textbf{0.004}\\
    &\SANA(AT) & \textbf{88.25} & \textbf{0.051} & \textbf{100} & \textbf{0.0538} & \textbf{82.66} & 0.017 & \textbf{66.37} & 0.0129 & \textbf{70.52} & \textbf{0.014} & \textbf{96.99} & 0.019\\
    \midrule
    \multirow{2}{*}{\textit{WINDOW}}
    &AE \cite{Comuzzi_Autoencoders}& 16.76 & 0.345 & 29.83 & 0.4708 & 20.05 & \textbf{0.0043} & 66.18 & \textbf{0.0035} & 27.75 & 0.0704 & 51.14 & \textbf{0.005}\\
    &\SANA(AT) & \textbf{79.06} & \textbf{0.062} & \textbf{66.6} & \textbf{0.0155} & \textbf{78.77} & 0.0325 & \textbf{68.88} & 0.026 & \textbf{58.5}2 & \textbf{0.018} & \textbf{95.04} & 0.044\\
    \midrule
    \multirow{2}{*}{\textit{RANDOM}}
    & AE \cite{Comuzzi_Autoencoders} & \textbf{66.44} & 0.973 & \textbf{79.12} & 1.703 & 28.96 &\textbf{0.0053} & \textbf{76.01} & \textbf{0.0034} & 47.48 & 0.106 & \textbf{91.3} & \textbf{0.006}\\
    & \SANA(AT) & 64.98 & \textbf{0.652} & 70.1 & \textbf{0.8944} & \textbf{58.33} & 0.1677 & 56.73 & 0.042 & \textbf{49.29} & \textbf{0.048} & 75.2 & 0.111\\
    \bottomrule
\end{tabular}
    }
    \vspace{1mm}
    \caption{Accuracy and Mean Absolute Error of missing events under different masking strategies (ODD, EVEN, WINDOW, RANDOM).}
    \vspace{-0.5cm}
    \label{tab:res_all_masks}
\end{table*}

\begin{table*}[t]
%
\begin{minipage}[t]{0.55\textwidth}
\scalebox{0.82}{
\begin{tabular}{lccccccccc
}
\toprule    
\multirow{2}{*}{\textbf{Masking}} & \multicolumn{2}{c}{\textbf{BPI12}} & \multicolumn{2}{c}{\textbf{BPI13}} & \multicolumn{2}{c}{\textbf{SP2020}} & \multicolumn{2}{c}{\textbf{BPI20RFP}}   \\  \cmidrule{2-9}
& Activity & Time & Activity & Time  & Activity & Time  & Activity & Time   \\ \toprule 

\textit{ODD} & $+3.29$ & $+0.021$ & $+9.98$ & $+0.465$ & $+9.61$ & $-0.003$ & $-1.45$ & $+0.069$ \\
\textit{EVEN} & $+2.16$ & $+0.021$ & $-1.57$ & $+0.477$ & $+5.65$ & $-0.003$ & $-0.08$ & $+0.068$ \\
\textit{WINDOW} &  $-8.96$ & $+0.043$ & $-9.32$ & $+0.374$ & $+2.2$ & $+0.008$ & $-0.17$ & $+0.077$\\
\textit{RANDOM} & $-4.27$ & $+0.094$ & $-3.29$ & $+0.388$ & $+1.62$ & $-0.006$ & $-6.16$ & $+0.071$\\
\bottomrule
\end{tabular}
}
\vspace{1mm}
\caption{Delta between \textit{FULL} and \textit{AT} configurations}
\vspace{-0.9cm}
\label{tab:res_delta_AT_FULL}
\end{minipage}
\quad
\begin{minipage}[t]{0.41\textwidth}
\scalebox{0.82}{
\begin{tabular}{lccccc}
\toprule
\multirow{2}{*}{\textbf{Method}} & \multirow{2}{*}{\textbf{Attribute}} & \multirow{2}{*}{\textit{ODD}} & \multirow{2}{*}{\textit{EVEN}} & \multirow{2}{*}{\textit{WINDOW}} & \multirow{2}{*}{\textit{RANDOM}} \\
& & & & & \\
\midrule
\SANA & Activity & 75.70 & 88.25 & 79.06 & 64.98 \\
2 layers & Time & \textbf{0.042 }& \textbf{0.051} & \textbf{0.062 }& 0.652 \\
\midrule
\SANA & Activity & \textbf{93.80} & \textbf{95.32} &\textbf{ 82.91} &\textbf{ 73.44} \\
4 layers & Time & 0.086 & 0.099 & 0.083 & \textbf{0.465 }\\
\bottomrule
\end{tabular}
}
\vspace{1mm}
\caption{Influence of network depth on the \textbf{Small Log}}
\vspace{-0.9cm}
\label{tab:res_layers}
\end{minipage}
\end{table*}

\textbf{RQ1}. Table~\ref{tab:res_all_masks} reports the results of \SANA and the ones of the Autoencoder approach from \cite{Comuzzi_Autoencoders}  (AE).
The table is organized according to the four experiments, each corresponding to one of the masking strategies described in the previous section. For each experiment, we highlight in bold the best-performing model.
%
For this comparison we use a version of our model that considers only event activities and timestamps. We denote this version as \SANA (AT).
This choice guarantees fairness in the comparison with \cite{Comuzzi_Autoencoders} since their model also considers only activity and timestamp as input, along with a computed attribute that accounts for \emph{accumulated time} from the first event in the event log, and it is capable of reconstructing only these two features.

Regarding activity reconstruction accuracy, we can observe that in the first three masking experiments, our model outperforms AE \emph{in all event logs}, with the single exception of the BPI13 log in the \textit{ODD} masking experiment.
Additionally, the wins are usually by a large margin (on average the difference in terms of accuracy is around 43 points).
Conversely, in the \textit{RANDOM} experiments AE results as the winning method for activity reconstruction, achieving a higher accuracy in 4 out of 6 event logs. It is worth noticing, however, that \SANA stands often behind by little margin, with the exception of the BPI12 log.
The reason for this behavior lies in the different learning paradigms of AE and \SANA.
AE reconstructs the entire trace by encoding global information into a latent space, using the full input regardless of where missing elements occur. It also reconstructs all activities, not just the missing ones, making it less sensitive to masking patterns.
\SANA, instead, builds hidden representations for each node based only on its neighbors within a limited receptive field defined by the convolutional layers. Thus, information propagation depends on the number of layers, limiting the context available for reconstruction.
As explained in Section~\ref{sec:Encoding}, when a sequence of empty nodes is longer than twice the number of convolutional layers, those nodes receive no useful information, rendering their representations ineffective. The \textit{RANDOM} masking experiment is the only one in which this situation may arise, since in the other experiments the maximum number of missing consecutive events is two (see the experiment reported in  the next section).

Regarding timestamp reconstruction the results of the two approaches are closer.
In fact for all four masking experiments each method wins on three event logs and loses on the other three: AE always wins in BPI12, BPI13 and BPI20RFP logs (with an average difference in terms of MAE of around 0.04), while \SANA(AT) always wins in the SP2020, Small Log and Large Log (with a average difference in terms of MAE of around 0.52).

\textbf{RQ2.} Table \ref{tab:res_delta_AT_FULL} reports the difference between the version of \SANA leveraging all event attributes to complete the missing information (FULL) and the \SANA version using only the information related to activity and timestamps, i.e., \SANA(AT), over the four real-life logs. \footnote{Results related to synthetic datasets are omitted as they only contain activity and timestamp attributes (as shown in Table \ref{tab:datasets}).} 
While, on the one hand, we might foresee better performance for FULL, as the model can leverage contextual
information from non-missing attributes, on the other hand, we expect a higher level of complexity, due to the multiple missing attribute
values it has to manage simultaneously.

By looking at the table we can observe that in the case of activity reconstruction, \SANA (FULL) is able to improve or only slightly degrade the \SANA(AT) performance. Indeed, although  the FULL version improves the accuracy of \SANA (AT) only in 7 out of 16 cases, the average difference between the two versions is of around -0.05. Focusing on the timestamps, we can observe that the AT version performs in general better than \SANA (FULL), except for the SP2020 log, in which FULL MAE is always lower. However, again, the average difference in terms of MAE is very low: \SANA(FULL) worsen \SANA(AT) of only 0.135 points. These considerations allow us to conclude that taking into account all attribute information does not worsen the \SANA(AT) results on activity and timestamp reconstruction (\ref{RQ2}).


\subsection{Discussion and  limitations}
\begin{figure*}[t]
    \centering
    \begin{minipage}[t]{0.45\textwidth}
        \centering
        \includegraphics[width=0.91\textwidth]{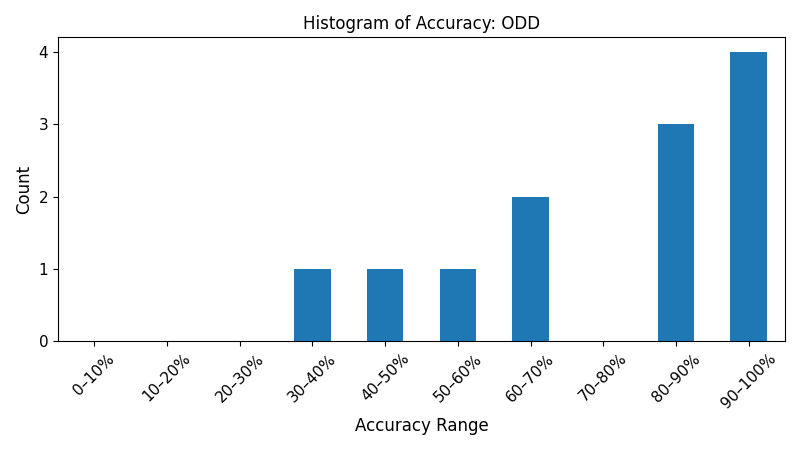}
    \end{minipage}\quad
    \begin{minipage}[t]{0.45\textwidth}
        \centering
        \includegraphics[width=0.91\textwidth]{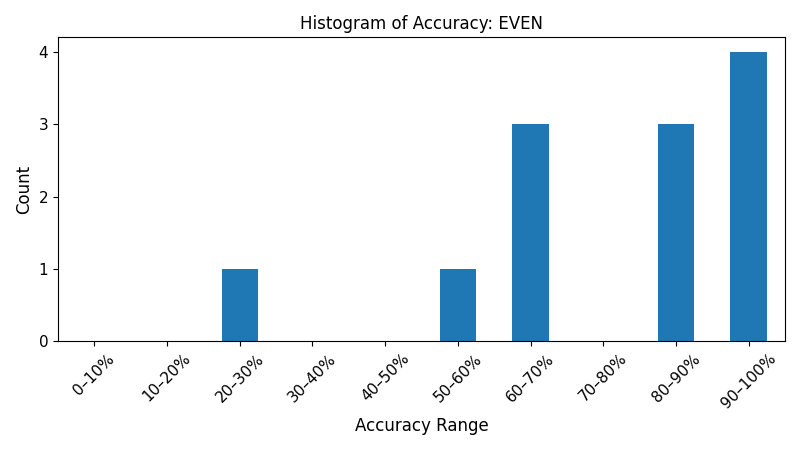}
    \end{minipage}
    \vspace{-.5cm}
    \caption{Distribution of reconstruction accuracy for categorical event attributes across the four real-world event logs, for the \textit{ODD} and \textit{EVEN}  experiments.}
    \label{fig:otherattributes}
\end{figure*}

\textbf{Other attributes.} Although a comparison of \SANA with state-of-the-art model-free approaches on the accuracy of reconstructing attributes different from activities and timestamps is not feasible, we still analyzed the performance of the approach on these attributes. The complete results on all attributes are available in the repository\footnote{\url{https://github.com/sebdisdv/SANAGRAPH/tree/main/repair\_results}}.
For space reasons, we provide here an overview in terms of accuracy on categorical event attributes for the \textit{ODD} and \textit{EVEN} masking experiments. Out of the 26 categorical attributes across the four real-world logs ($\sim\!90\%$ of the total number of attributes), we report in Figure~\ref{fig:otherattributes} the distribution of the reconstruction accuracy on the 12 categorical event attributes. The plots show that, both for the \textit{ODD} and \textit{EVEN} masking strategy, although for 2 or 3 attributes the accuracy is lower than 50\%, the majority (more than half) reach an accuracy higher than 80\%.

\label{subsec:model_depth}
\textbf{Limitations.} A limitation of our approach, as introduced in Section \ref{sec:Encoding}, concerns the choice of the number of layers in our model. To evaluate the importance of the number of layers, we conducted a small experiment on the Small Log dataset, where we doubled the number of layers up to 4, and compared it to the 2 layer setting. The results are reported in Table \ref{tab:res_layers}. We can see an overall increase in the accuracy on the activity label, and a smaller MAE in the \textit{RANDOM} dataset. From this, we can estimate that by increasing the number of layers, we can indeed achieve better performance. However, according to the GNN literature \cite{Number_of_layers}, there exists a threshold number of convolutional layers, beyond which, the nodes would already contain all the available information in the entire graph, so that further increasing the network depth would be to no purpose. We leave the study of this phenomenon, as well as the possibility to include the number of layers in the hyperparameter search, for future work.


\section{Conclusion}
\label{sec:END}
In this work, we propose a new method to tackle the event log repair problem in case of missing information. Our method involves the use of GNNs, more specifically HGNNs, which allow for a richer encoding of event log traces. \SANA does not only reconstruct the main attributes of an event, such as the activity or the timestamp, but it is also able to reconstruct all other attributes at the same time, which, to the best of our knowledge, no other DL repair methods available in the literature is able to do. The comparison with state-of-the-art approaches shows higher accuracy when reconstructing missing activities and a MAE in line with the state-of-the-art when reconstructing missing attributes. Moreover, it also reaches good performance across all the other attributes. 
Future directions would include a richer encoding of the graph in input, by leveraging global features that can be extracted from the event log. Moreover, we plan to provide some form of explanation together with the reconstructed trace.

\section*{Acknowledgment}
This work has been partially funded by the PNRR project FAIR - Future AI Research
(PE00000013), under the NRRP MUR program funded by the NextGenerationEU.

\bibliographystyle{ieeetr}

\end{document}